\documentclass{article}

\usepackage[preprint, nonatbib]{neurips_2022}

\usepackage[utf8]{inputenc} % allow utf-8 input
\usepackage[T1]{fontenc}    % use 8-bit T1 fonts
\usepackage{hyperref}       % hyperlinks
\usepackage{url}            % simple URL typesetting
\usepackage{booktabs}       % professional-quality tables
\usepackage{amsfonts}       % blackboard math symbols
\usepackage{nicefrac}       % compact symbols for 1/2, etc.
\usepackage{microtype}      % microtypography
\usepackage[dvipsnames]{xcolor}
\usepackage{amsthm}
\usepackage{amsmath}
\usepackage{graphicx}
\usepackage{color, colortbl}
\usepackage{wrapfig}
\usepackage{pifont}
\usepackage{amssymb}

\definecolor{LightCyan}{rgb}{0.88,1,1}

\title{Masked Siamese ConvNets}

\author{
Li Jing\thanks{Equal Contribution} \\
Meta AI \\
\texttt{ljng@fb.com}
\And
Jiachen Zhu$^*$ \\
NYU \\
\texttt{jiachen.zhu@nyu.edu}
\And
Yann LeCun \\
Meta AI \& NYU \\
\texttt{yann@fb.com}
}

\begin{document}

\maketitle

\begin{abstract}
Self-supervised learning has shown superior performances over supervised methods on various vision benchmarks. The siamese network, which encourages embeddings to be invariant to distortions, is one of the most successful self-supervised visual representation learning approaches. Among all the augmentation methods, masking is the most general and straightforward method that has the potential to be applied to all kinds of input and requires the least amount of domain knowledge. However, masked siamese networks require particular inductive bias and practically only work well with Vision Transformers. This work empirically studies the problems behind masked siamese networks with ConvNets. We propose several empirical designs to overcome these problems gradually. Our method performs competitively on low-shot image classification and outperforms previous methods on object detection benchmarks. We discuss several remaining issues and hope this work can provide useful data points for future general-purpose self-supervised learning.
\end{abstract}

\section{Introduction}
\label{sec:intro}

Self-supervised learning aims to learn useful representations from scalable unlabeled data without relying on human annotation. It has succeeded in natural language processing~\cite{Devlin2019BERTPO, Zhang2022OPTOP, Brown2020LanguageMA}, speech recognition~\cite{Oord2018RepresentationLW, Hsu2021HuBERTSS, Schneider2019wav2vecUP, Baevski2020wav2vec2A} and other domains~\cite{Rives2021BiologicalSA, Rong2020SelfSupervisedGT}. Self-supervised visual representation learning has also become an active research area. 

First introduced in ~\cite{Bromley1993SignatureVU}, the siamese network~\cite{chen2020simple, Chen2020BigSM, He2020MomentumCF, Chen2020ImprovedBW, Chen2021AnES, caron2020swav, grill2020byol, chen2020simsiam, Wang2022OnTI, zbontar2021barlow, Bardes2021VICRegVR} is one promising approach among many self-supervised learning approaches and outperforms supervised counterparts in many visual benchmarks. It encourages the encoder to be invariant to human-designed augmentations, capturing only the essential features.
Practically, the siamese network methods rely on domain-specific augmentations, such as cropping, color jittering~\cite{Wu2018UnsupervisedFL} and Gaussian blur~\cite{chen2020simple}, which do not apply to new domains. Therefore, it is desirable to find a general augmentation approach that requires minimal domain knowledge.

Among various augmentations, masking or disrupting the input remains one of the simplest and most effective methods, which has been demonstrated to be useful for NLP~\cite{Devlin2019BERTPO} and speech~\cite{Hsu2021HuBERTSS}. However, not until the recent success of vision transformers (ViTs)~\cite{Dosovitskiy2021AnII, Touvron2021TrainingDI} can vision models leverage masking as a general augmentation. Self-supervised learning with masking has demonstrated more scalable properties when combined with ViTs~\cite{He2021MaskedAA, Bao2021BEiTBP, Zhou2021iBOTIB, Baevski2022data2vecAG}. Unfortunately, siamese networks with naive masking do not work well with most off-the-shelf architecture, e.g., ConvNets~\cite{He2016DeepRL, Liu2022ACF}.

\begin{figure}[t!]
    \centering
    \includegraphics[width=\textwidth]{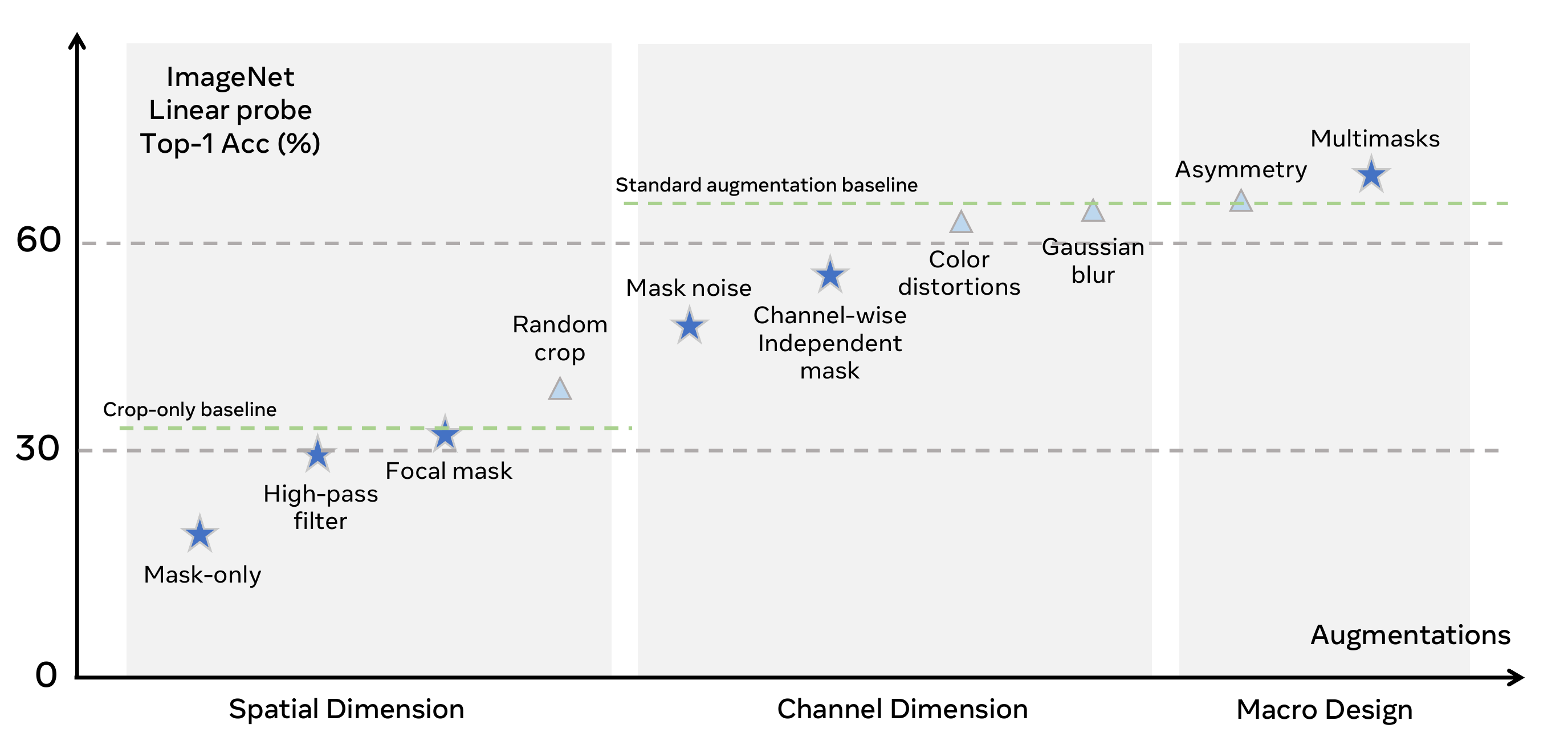}
    \caption{\textbf{Masking Design}. We propose a series of designs for masked siamese networks with ConvNets. The stars represent our masking design, and the triangles represent standard augmentations applied to the original image. Our masking design spans spatial dimension, channel dimension, and macro design. The pretraining is conducted on ImageNet-1K training set for 100 epochs using ResNet-50 backbone. Each model is evaluated on the validation set using frozen features with a linear classifier. The final model achieves $67.6\%$ accuracy and outperforms the unmasked baseline by $1.0\%$.}
    \label{fig:main_results}
\end{figure}

This work identifies the underlying issues behind masked siamese networks with ConvNets. We argue that masked inputs create parasitic edges, introduce superficial solutions, distort the balance between local and global features, and have fewer training signals. We propose several designs to overcome these problems gradually. See Figure~\ref{fig:main_results}. Experiments show that siamese networks with ConvNets backbone can benefit from masked inputs with these designs.

We discuss the current challenge and open questions about siamese networks with masked inputs. We hope this work will provide useful insights for future general-purpose self-supervised learning with minimal domain knowledge or architecture inductive bias.

We summarize our contributions below:
\begin{itemize}
    \item We identify the underlying problems why masked siamese networks perform poorly with ConvNets backbones.
    \item We propose several empirical designs and gradually overcome these problems for masked siamese networks with ConvNets.
    \item We propose Masked Siamese ConvNets (MSCN), which performs competitively on low-shot image classification benchmarks and outperforms previous methods on object detection benchmarks.
\end{itemize}

\section{Related Works}
\label{sec:related}

\subsection{Siamese Networks}

Self-supervised visual representation learning has become an active research area since they have shown superior performances over supervised counterparts in recent years. One promising approach is to learn useful representations by encouraging them to be invariant to augmentations, known as siamese networks or joint-embedding methods~\cite{Misra2020SelfSupervisedLO, chen2020simple, Chen2020BigSM, He2020MomentumCF, Chen2020ImprovedBW, Chen2021AnES, caron2020swav, grill2020byol, chen2020simsiam, Wang2022OnTI, zbontar2021barlow, Bardes2021VICRegVR}. These methods use different mechanisms to prevent collapse, and they all rely on carefully designed augmentations such as random resized cropping, color jittering, grayscale and Gaussian blur. These augmentations prevent the encoder from only using trivial features. Empirically, siamese networks with these standard augmentation settings usually work well with off-the-shelf architectures, including ResNets~\cite{He2016DeepRL} and ViTs~\cite{Dosovitskiy2021AnII}. Their representations are label-efficient~\cite{Assran2021SemiSupervisedLO, Assran2022MaskedSN}, more robust~\cite{Hendrycks2019UsingSL}, and have improved fairness~\cite{Goyal2022VisionMA}. In addition,  Siamese networks have been demonstrated to benefit from scalable data~\cite{Goyal2021SelfsupervisedPO}.

\subsection{Representation learning with Masked Images}

Masking or corrupting the input is one of the simplest augmentations and could be applied to a wide range of data types. Masking is mostly used in the transformer-based denoising autoencoder frameworks~\cite{Vincent2010StackedDA, Vincent2008ExtractingAC}. Motivated by the success in NLP~\cite{Devlin2019BERTPO, Brown2020LanguageMA} with transformers~\cite{Vaswani2017AttentionIA}, many visual representation learning methods~\cite{He2021MaskedAA, Bao2021BEiTBP, Zhou2021iBOTIB} using ViTs have also shown benefit from masked inputs. These methods have demonstrated improved finetuning performance and good scalability.

This successful combination of masked images and ViTs is also helpful for siamese networks~\cite{Baevski2022data2vecAG, Assran2021SemiSupervisedLO}, where masking becomes an extra augmentation with the benefits of significantly improved efficiency. However, no previous work has shown that the masking approach can work equally well with off-the-shelf ConvNets.

% \subsection{Architectures Inductive Bias}

% ConvNets have dominated computer vision ever since AlexNet~\cite{Krizhevsky2012ImageNetCW} won the ImageNet~\cite{Krizhevsky2012ImageNetCW} competition by a big margin. Recently, vision transformers~\cite{Dosovitskiy2021AnII} revolutionized the architecture design by taking advantage of the transformers~\cite{Vaswani2017AttentionIA} brought from NLP domains. Many new architecture designs~\cite{Tolstikhin2021MLPMixerAA} do not require a convolutional inductive bias. Later, it has been shown that  shows that with careful micro and macro designs, ConvNets can also perform competitive with ViTs on supervised learning scenarios~\cite{Liu2022ACF}.

% Given that there's a fundamental differences between ConvNets' and ViTs' inductive bias, it is desirable to look for training algorithms that work equally well regardless of the architecture. 

\section{Problems in Masked Siamese Networks with ConvNets}
\label{sec:problem}

Siamese networks with masked inputs have demonstrated competitive performances~\cite{Assran2022MaskedSN, Baevski2022data2vecAG} using ViTs. Naively replacing ViTs with off-the-shelf ConvNets results in significantly worse performances.  Here, we identify the underlying problems.

\textbf{Masking Introduces Parasitic Edges} -
The convolutional kernels are well known for their edge detection behaviors~\cite{LeCun1998GradientbasedLA}. Applying masks creates a large number of parasitic edges in the images. The feature maps generated by edge detecting kernels are drastically distorted, and hence these kernels are suppressed during training. More severally, these parasitic edges will remain in the output feature maps and affect all the hidden layers. On the contrary, ViTs dodged this problem because the mask is usually designed to match the patch boundaries. 

\begin{wrapfigure}{r}{0.45\textwidth}
\vspace{-0.2in}
  \begin{center}
    \includegraphics[width=0.2\textwidth]{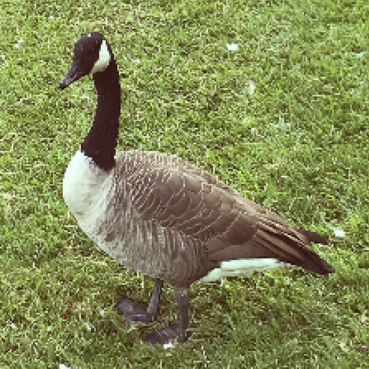}
    \includegraphics[width=0.2\textwidth]{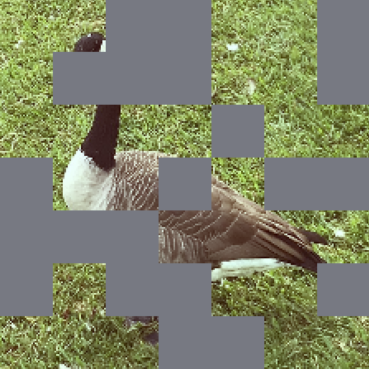} \\
    \includegraphics[width=0.2\textwidth]{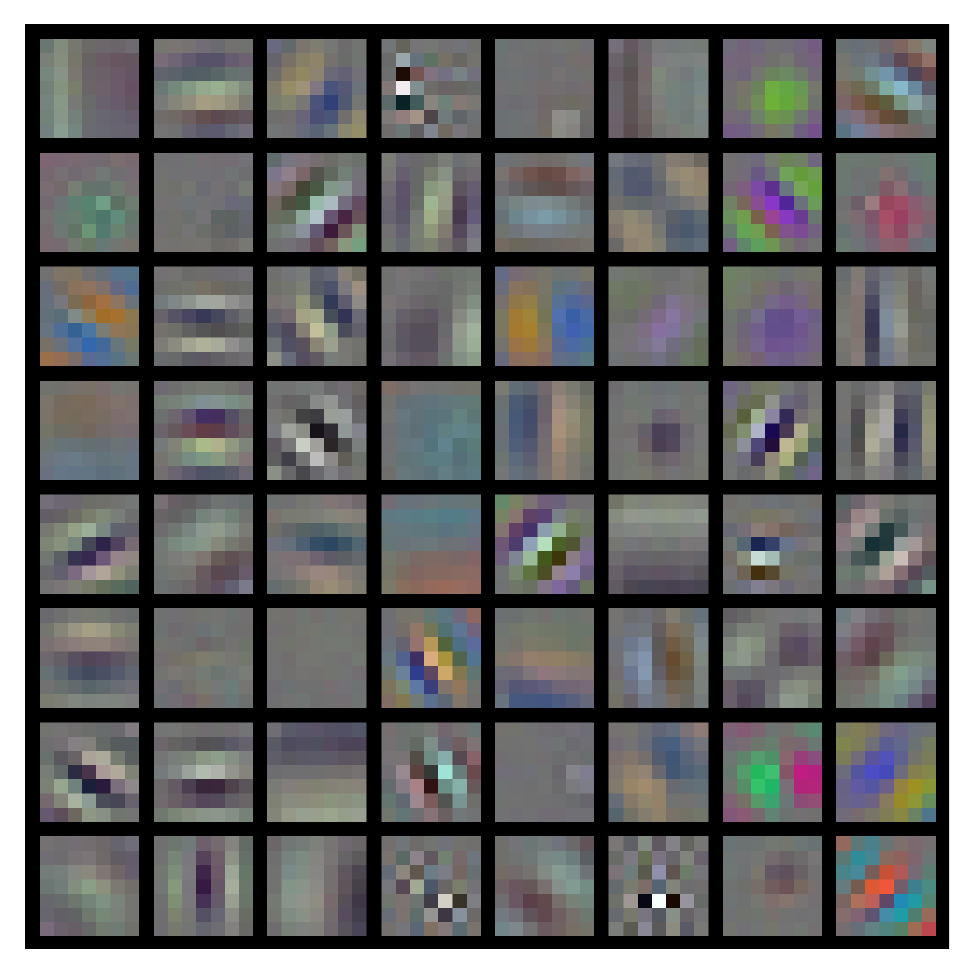}
    \includegraphics[width=0.2\textwidth]{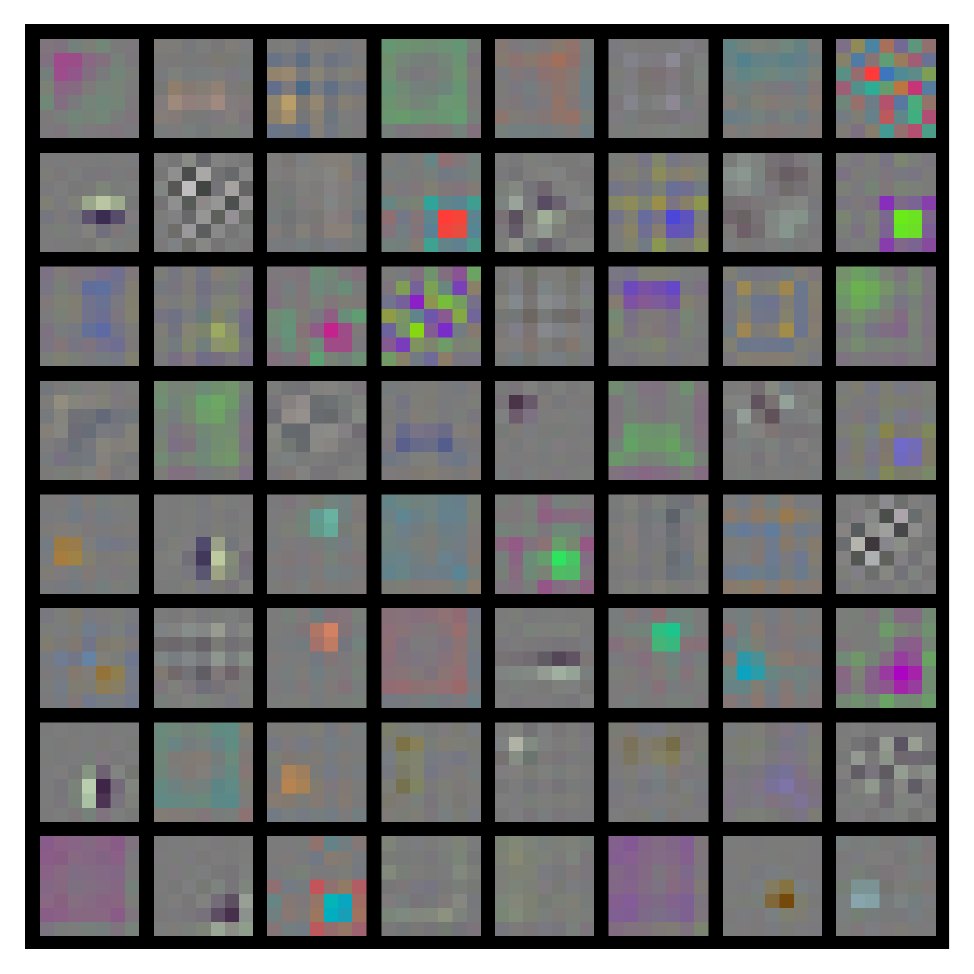}
    \caption{Visualization of the encoder's first convolutional layer kernels pretrained with different augmentation settings. Left: with standard augmentation, the kernels show normal edge detection behaviors. Right: with masking, a large number of kernels collapse and the network is unable to detect useful features.}
    \label{fig:kernels}
    \end{center}
\vspace{-0.2in}
\end{wrapfigure}

In Figure~\ref{fig:kernels}, we visualize the encoder’s first
convolutional layer kernels pretrained with standard augmentation or masked input. Due to the parasitic edges, many kernels collapsed to trivial blank features.

\textbf{Masking Introduces Superficial Solutions} - Standard augmentations in siamese networks can prevent the network from only leveraging superficial solutions. For example, using a color histogram as representation becomes infeasible with color jittering. Naively applying masks will introduce new superficial solutions such that the network can leverage features only based on the masked area. For example, the mask shape might be a trivial feature. Furthermore, with symmetric filling, the masked area contains the same color histogram and similar patterns. ViTs also avoid this problem as the network will only focus on the unmasked areas.

\textbf{Balance between Local and Global Features} - Random resized cropping is the most critical augmentation for siamese networks. By varying the scale of the crops, the siamese networks find a precise combination of short-range and long-range correlation, known as local/textural features and global/semantic features. Cropping can be considered a special case of masking, but random masking distorts local and global features with a different ratio based on mask grid size. In ViTs, the mask grid size is fixed and is set to match the patch size. Therefore, a spatial mask design makes little difference to this balance for ViTs. However, ConvNets with scale invariance inductive bias may benefit from a careful spatial masking design.

\textbf{Less Learning Signal} - Masked inputs only contain partial information, which results in less learning signal. Practically, masked approaches usually need to train significantly longer or use multicrops~\cite{caron2020swav, Caron2021EmergingPI}. For example, masked autoencoder~\cite{He2021MaskedAA} benefits from longer training up to 1600 epochs. Masked siamese networks~\cite{Assran2022MaskedSN} improves ImageNet-1K accuracy by more than $20\%$ by using extra multicrops. This also results in less computation efficiency for ConvNets as they cannot skip the unmasked areas like ViTs.

\section{Designing Masked Siamese ConvNets}
\label{sec:design}

In this section, we propose several empirical designs to overcome the problems discussed in Sec~\ref{sec:problem} and show a trajectory to our final masking strategy. We use SimCLR~\cite{chen2020simple} with ResNet-50 backbone~\cite{He2016DeepRL} as our study environment. For experiments in this section, we pretrain each model for 100 epochs on ImageNet-1K~\cite{Deng2009ImageNetAL} training set using LARS optimizer~\cite{You2017LargeBT} with a batch size of 4096. All the results are linear probe accuracy on ImageNet-1K validation set. The evaluation detail is in the supplementary material.

\subsection{Preliminaries}

The goal of siamese networks is to learn representations of input images so that they can be used for downstream tasks. Most methods start with randomly creating two crops $\textbf{x}_1$ and $\textbf{x}_2$ from the same input $\textbf{x}$, and applying twos sets of randomly augmentation transformations $T_{\phi}$ and $T_{\phi'}$ to the crops. The siamese networks then train an encoder $f_{\theta}(\cdot)$ so that $||f_{\theta}(T_{\phi}(\textbf{x}_1)) - f_{\theta}(T_{\phi'}(\textbf{x}_2))||^2 \rightarrow 0$, $\forall \textbf{x}$ and $\forall \phi$. This is known as the positive term in siamese networks.

Suppose we only train the encoder $f_{\theta}(\cdot)$ with the positive term alone. In that case, it could quickly converge to the collapse solution that produces a constant representation for every input $\textbf{x}$.
Preventing collapse is solved by various frameworks, including contrastive loss~\cite{chen2020simple, He2020MomentumCF}, redundancy reduction~\cite{zbontar2021barlow, Bardes2021VICRegVR}, clustering~\cite{caron2020swav} and distillation~\cite{grill2020byol, chen2020simsiam, Caron2021EmergingPI}. These methods guarantee that $\mathbb{E}_{\phi, \phi'}[||f_{\theta}(T_{\phi}(\textbf{x})) - f_{\theta}(T_{\phi'}(\textbf{x}'))||^2] > \epsilon $ with a hyperparameter $\epsilon$ for $\textbf{x}, \textbf{x}'$ coming from different images.  This is known as the negative term in siamese networks.

In this work, the positive term and the augmentations $T_{\phi}$ are our primary focus. Proper designed $T_{\phi}$ is essential for learning good representations since siamese networks without it do not guarantee that all features in $f_{\theta}(\cdot)$ are useful for downstream tasks.
Consider a useful feature $f$ and a trivial feature $g$ respect to a given task, both of which satisfies negative term. The siamese network can benefit from using augmentation $T_{\phi}$ if $||f(T_{\phi}(\textbf{x}_1)) - f(T_{\phi'}(\textbf{x}_2))||^2 \approx ||f(\textbf{x}_1) - f(\textbf{x}_2)||^2$ and $||g(T_{\phi}(\textbf{x}_1)) - g(T_{\phi'}(\textbf{x}_2))||^2 \gg ||g(\textbf{x}_1) - g(\textbf{x}_2)||^2$. Since $g$ will lead to a higher positive term, then the encoder $f_\theta$ is more likely converge to $f$ instead of $g$ through training.
So we remove the trivial features from the representation by adding a data augmentation to the pretraining pipeline.

Moreover, the sub-optimal performance of traditional pattern recognition with handcrafted features on image classification or object detection indicates that useful features for those tasks do not have mathematical or conceptual simplicity. So, when we design our augmentations, we are looking for mathematical or conceptual simple features and coming up with augmentation to prevent the network from converging to such features.

\subsection{Designing Principle}

Standard augmentations prevent superficial features based on simple input statistics. However, with masked inputs, superficial features may leverage the masked areas and surpass useful ones. 

We denote the mask as $\textbf{M}$ and the filling values for the masks area as $\textbf{z}$. This masked image can be written as $\textbf{M}*\textbf{x} + (1-\textbf{M}) * \textbf{z}$. Therefore, we derive our masking design principle. For a useful feature $f$ and a trivial feature $g$, we require $\textbf{M}$ and $\textbf{z}$ to satisfy:   $||f(\textbf{M}*\textbf{x} + (1-\textbf{M}) * \textbf{z}) - f(\textbf{M}'*\textbf{x} + (1-\textbf{M}') * \textbf{z}')||^2 \approx ||f(\textbf{x}_1) - f(\textbf{x}_2)||^2$ and $||g(\textbf{M}*\textbf{x} + (1-\textbf{M}) * \textbf{z}) - g(\textbf{M}'*\textbf{x} + (1-\textbf{M}') * \textbf{z}')||^2 \gg ||g(\textbf{x}_1) - g(\textbf{x}_2)||^2$.

Next, we start to build masking strategy following our designing principle.

\subsection{Spatial Dimension}

We first focus on spatial dimension to study how to best leverage masking in siamese networks. 

We start with applying two random grid masks (grid size 32) on the same random crop with a fixed $30\%$ masking ratio and no other augmentations. This mask-only setting achieves a non-trivial $\mathbf{21.0\%}$.

\begin{figure}[h!]
    \centering
    \includegraphics[width=0.2\textwidth]{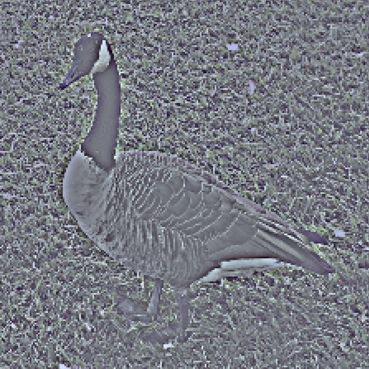}
    \includegraphics[width=0.2\textwidth]{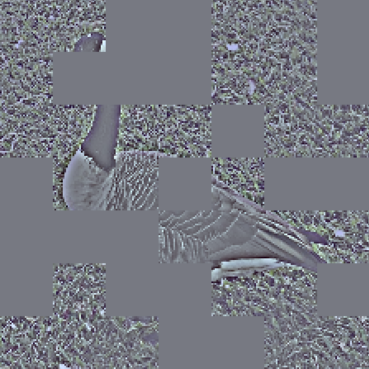}
    \includegraphics[width=0.2\textwidth]{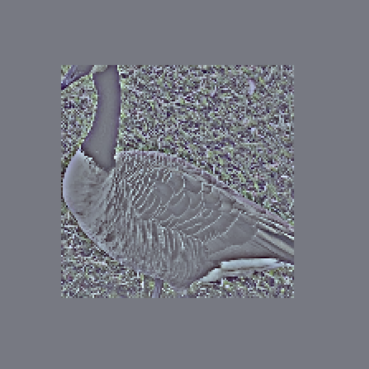}
    \caption{\textbf{Spatial Dimension design}. From left to right:  image with a \textit{High-pass Filter}, \textit{Random Grid Mask}, \textit{Focal Mask}. The high-pass filter shifts the pixel value distribution so that zeros represent null information. After applying the high-pass filter, the mask edges become invisible. A focal mask is a special case of a random grid mask and can also be considered as cropping without resizing. Random grid mask and focal mask distorts local and global features differently.}
    \label{fig:design-spatial}
\end{figure}

To overcome the problem of the parasitic edges introduced by arbitrary grid mask boundaries, we apply a high-pass filter before applying masks. See Figure~\ref{fig:design-spatial}. With a high-pass filter, the parasitic edges become invisible. In addition, the special value $0$ in the input image represents null information instead of normal pixel values. We find that $\sigma=5$ is optimal. With a high-pass filter, the model accuracy increases to $\mathbf{30.2\%}$.

As discussed in Sec~\ref{sec:problem}, it is crucial to balance short-range and long-range features in the input to learn useful representations. Following \cite{Assran2022MaskedSN}, we apply focal masks in addition to random grid masks. See Figure~\ref{fig:design-spatial}. The focal masks can be seen as a random crop without resizing. We apply a focal mask with $20\%$ probability and a grid mask with $80\%$. Different from \cite{Assran2021SemiSupervisedLO}, our combination of random grid mask and focal mask samples randomly. This improves the model accuracy to $\mathbf{31.0\%}$.

Lastly, we combine our spatial masking design with standard random resized cropping. We allow the two branches to use different cropped views. This combined approach achieves $\mathbf{40.0\%}$ accuracy. Note that without masking, a model using crop-only augmentation only obtains $\mathbf{33.5\%}$ accuracy.

\textit{(High-pass Filter) - We will apply high-pass filter before applying mask.}

\textit{(Focal Mask) - We will randomly apply focal mask in addition to random grid mask.}

\subsection{Channel Dimension}

% Color-jittering and grayscale are important channel-dimension augmentations as they prevent the encoders from leveraging superficial features such as simple color histograms.

\begin{wrapfigure}{r}{0.45\textwidth}
\vspace{-0.2in}
  \begin{center}
    \includegraphics[width=0.2\textwidth]{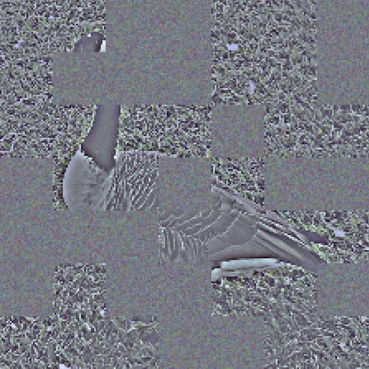}
    \includegraphics[width=0.2\textwidth]{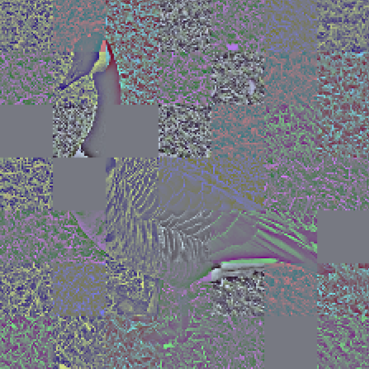}
    \caption{\textbf{Channel Dimension Design}. Left: \textit{Mask Noise}. We apply Gaussian noise to the masked area to distort the overall color histogram. Right: \textit{Channel-wise Independent Mask}. With $70\%$ probability, we apply three random masks on three color channels independently. This distorts the correlation between different color dimensions.}
    \label{fig:design-channel}
    \end{center}
\vspace{-0.2in}
\end{wrapfigure}

Now we focus on designing masks on channel dimensions. First, we find that adding noise to the masked area is beneficial. See Figure~\ref{fig:design-channel}. This prevents the network from taking advantage of the overall color histogram and is equivalent to applying color-jittering on the masked area. Adding noise to the masked area improves the accuracy from $40.0\%$ to $\mathbf{48.2\%}$.

Next, we randomly apply a channel-wise independent mask. In addition to standard spatial-wise masking, where we apply the same mask on three color channels, we generate three random masks and apply them to each color channel separately. We find that applying channel-wise independent masks with $70\%$ probability is optimal. See Figure~\ref{fig:design-channel}. This improves the accuracy to $\mathbf{53.6\%}$.

Lastly, We combine our channel-wise masking design with standard augmentations. By applying color-jittering and grayscale on the two branches before applying masks, the model achieves $\mathbf{63.0\%}$ accuracy. Next, randomly applying Gaussian blur on the two branches improves the accuracy to $\mathbf{65.1\%}$. 

\textit{(Mask Noise) - We will apply noise values in the masked area with varying standard deviation on different branches.}

\textit{(Channel-wise Independent Mask) - We will apply channel-wise independent mask with $70\%$ probability in addition to spatial-wise mask .}

\subsection{Macro Designs}

As suggested by~\cite{Wang2022OnTI}, we find that increasing asymmetry between the two networks improves accuracy. By varying the probability between two branches, the model accuracy improves to $\mathbf{65.6\%}$.

As discussed in Sec~\ref{sec:problem}, masked siamese networks receive less information in each iteration. We generate multiple masked inputs and apply joint-embedding loss on asymmetric pairs. This is similar to multicrops~\cite{caron2020swav} and benefits from amortized representations. This multimasks design increase accuracy to $\mathbf{67.4\%}$. Our final design is $\mathbf{1.0\%}$ better than not applying masks, and $\mathbf{5.2\%}$ better than using standard augmentation plus a naive random mask.

\textit{(Asymmetry) - We will apply augmentations asymmetrically on different branches.}

\textit{(Multimasks) - We will generate multiple masked versions.}

\subsection{Design Summary}

Following our design principle, we have gradually improved the masking strategy. We summarize the overall design below:
\begin{enumerate}
    \item apply standard augmentations: \\ RandomResizedCrop, HorizontalFlip, ColorJitter, Grayscale, GaussianBlur
    \item apply a high-pass filter
    \item apply masks \\
    spatial dimension: focal mask and random grid mask \\
    channel dimension: channel-wise independent mask and spatial-wise mask \\
    add random noise to the masked area
    \item increase asymmetry between different branches
    \item apply multimasks
\end{enumerate}

\begin{figure}[t!]
    \centering
    \includegraphics[width=\textwidth]{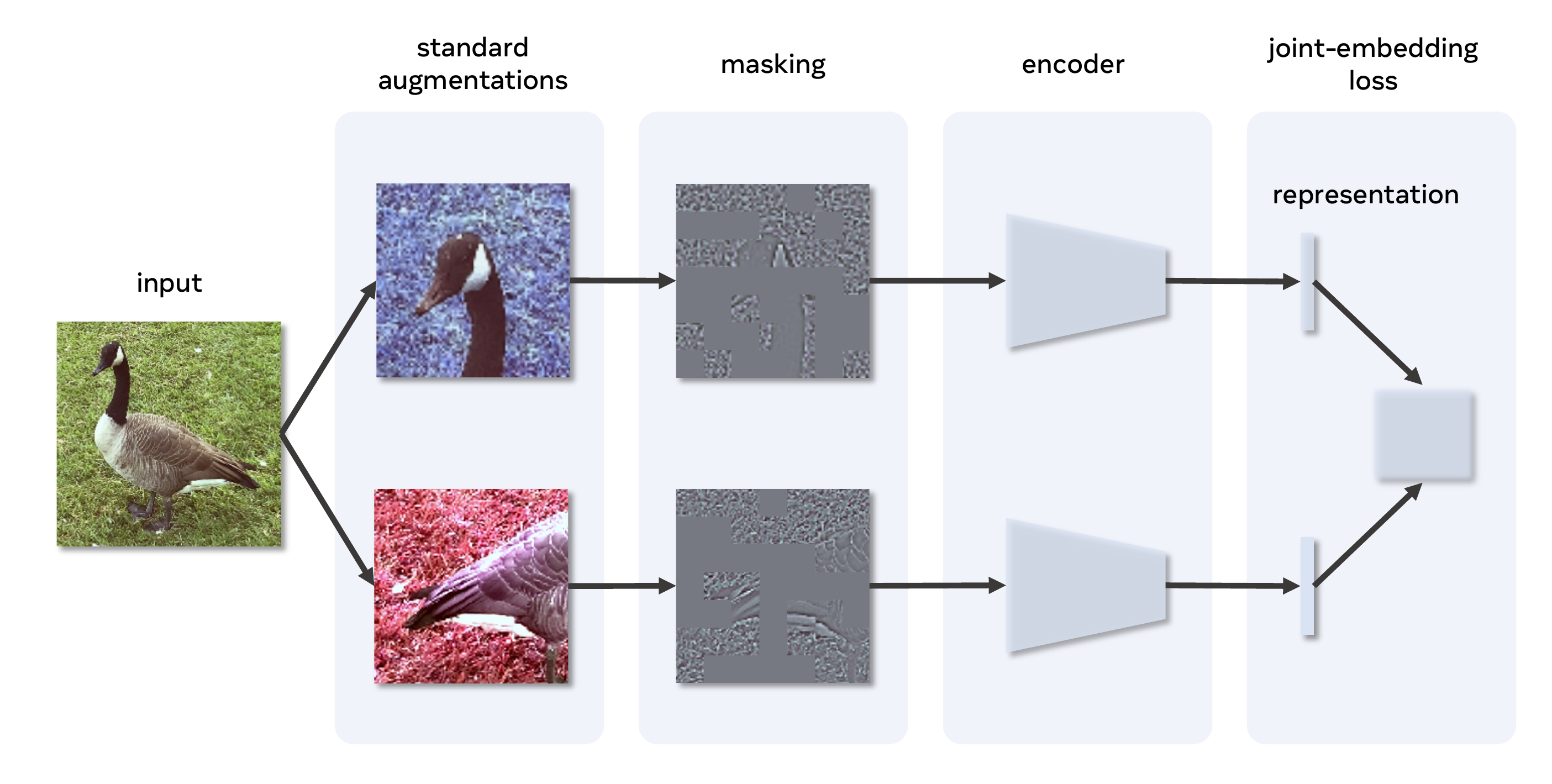}
    \caption{\textbf{Masked Siamese ConvNets (MSCN) framework}. MSCN first generates multiple views from the input image using a series of standard augmentations. Then it applies random masks on each view using our masking strategy. An off-the-shelf ConvNet encoder generates representations of these masked views. An off-the-shelf joint-embedding loss is applied to pairs of these representation vectors.}
    \label{fig:MSCN}
\end{figure}

The overall Masked Siamese ConvNets (MSCN) architecture is shown in Figure~\ref{fig:MSCN}. MSCN utilizes arbitrary backbone architecture and various joint-embedding loss functions.

\section{Results}
\label{sec:experiments}

In this section, we evaluate the representations obtained by a Masked Siamese ConvNets pretrained on the ImageNet-1K dataset~\cite{Russakovsky2015ImageNetLS} with ResNet-50 backbone~\cite{He2016DeepRL}. We train MSCN with different joint-embedding losses, namely SimCLR and BYOL. The network is pretrained for 800/1000 epochs, including 10 epochs of warm-up, and uses a cosine learning rate schedule. We use the LARS optimizer~\cite{You2017LargeBT} with a batch size of 4096. All the hyperparameters, including learning rate, closely follow the original SimCLR and BYOL implementation so that we could have a fair comparison.
For each model, the pretraining is distributed across 32 V100
GPUs and takes approximately 180 hours.
We list all the pretraining and evaluation details in supplementary material. 

\subsection{Image Classification}

We first evaluate the representations on the ImageNet-1K dataset using linear probe and semi-supervised classification. We compare MSCN with baselines in Table~\ref{tab:linear_probe_siamese}.

For linear probe, we train a linear classifier on $100\%$ of the labels with frozen weights. MSCN improves SimCLR but performs slightly worse with the BYOL baseline.
We suspect BYOL may require a different asymmetric masking setting due to its asymmetric nature of the framework.

For semi-supervised classification, we finetune the network using $1\%$ of the labels. MSCN demonstrates superior performances. The advantage of masked siamese networks in low-shot image classification tasks has also been observed in \cite{Assran2022MaskedSN} using ViTs.

We compare the effect of masking on ConvNets and ViTs in Table~\ref{tab:linear_probe_mask}.  MSCN with a ConvNet backbone demonstrates similar behaviors to MSN with a ViT backbone.

\begin{table}[h!]
    \centering
    \caption{\textbf{Evaluation on ImageNet-1K.} We evaluate the MSCN representations with linear probe and semi-supervised classification. For the linear probe, a linear classifier is trained on $100\%$ of the labels with frozen weights. For semi-supervised learning, we finetune the network with $1\%$ of the labels. MSCN significantly improves the SimCLR baseline but is slightly worse with BYOL. MSCN also demonstrates competitive performance on semi-supervised benchmarks in low-shot regions.}
    \begin{tabular}{lll}
    \textbf{Method} & \textbf{Linear Evaluation} & \textbf{Finetuning}\\
    Label fraction & $100\%$ & $1\%$ \\
    \toprule
    SimCLR~\cite{chen2020simple}  & 69.5 & 48.3\\
    \rowcolor{LightCyan} \ \ \ \  +MSCN & 71.5 \textcolor{SeaGreen}{\small (+2.0)} & 54.0  \textcolor{SeaGreen}{\small (+5.7)}\\
    \midrule
    BYOL~\cite{grill2020byol}  & 74.3 & 53.2 \\
    \rowcolor{LightCyan} \ \ \ \ +MSCN  & 74.4  \textcolor{SeaGreen}{\small (+0.1)} & 55.2  \textcolor{SeaGreen}{\small (+2.0)} \\
    % \midrule
    % VICReg~\cite{Bardes2021VICRegVR}  & 73.2 & 54.8 \\
    % \rowcolor{LightCyan} \ \ \ \ +MSCN  & ? & ? \\
    \bottomrule
    \end{tabular}
    \label{tab:linear_probe_siamese}
\end{table}

\begin{table}[h!]
    \centering
    \caption{\textbf{Effect of Masking on ConvNets and ViTs}. We compare the effect of masking on ConvNets and ViTs. MSCN with a ConvNet backbone demonstrates similar behaviors to MSN with a ViT backbone.}
    \begin{tabular}{llccc}
    \textbf{Method} & \textbf{Architecture} & \textbf{Parameters}  &\textbf{Use Mask} & \textbf{Accuracy}\\
    \toprule
    \textcolor{gray}{ Supervised}~\cite{Touvron2021TrainingDI} & \textcolor{gray}{ViT-S} & \textcolor{gray}{22M} & & \textcolor{gray}{79.9} \\
    \textcolor{gray}{MAE}~\cite{He2021MaskedAA} & \textcolor{gray}{ViT-S} & \textcolor{gray}{22M} & \textcolor{gray}{\ding{51}} & \textcolor{gray}{68.0} \\
    DINO~\cite{Caron2021EmergingPI} & ViT-S & 22M & \ding{55} & 78.3 \\
    MSN~\cite{Assran2022MaskedSN} & ViT-S & 22M & \ding{51} & 76.9 \\
    \midrule
    \textcolor{gray}{Supervised}~\cite{He2016DeepRL} & \textcolor{gray}{ResNet-50} & \textcolor{gray}{24M} & & \textcolor{gray}{76.5} \\
    % SimCLR~\cite{chen2020simple} & ResNet-50 & 24M & \ding{55} & 69.5 \\
    % VICReg~\cite{Bardes2021VICRegVR} & ResNet-50 & 24M & \ding{55} & 73.2 \\
    BYOL~\cite{grill2020byol} & ResNet-50 & 24M & \ding{55} & 74.3 \\
    % \rowcolor{LightCyan} MSCN (w/ SimCLR) & ResNet-50 & 24M & \ding{51} & 71.4\\
    \rowcolor{LightCyan} MSCN & ResNet-50 & 24M & \ding{51} & 74.4\\
    \bottomrule
    \end{tabular}
    \label{tab:linear_probe_mask}
\end{table}

\subsection{Transfer Learning}

We then evaluate the representations by transferring the network to other downstream tasks. 
We report the transferred image classification results on iNaturalist 2018~\cite{Horn2018TheIS} dataset and Places-205~\cite{Zhou2014LearningDF} dataset in Table~\ref{tab:transfer-image-classification}.
In Table~\ref{tab:transfer-object-detection}, we report the object detection and instance segmentation performance on VOC07+12~\cite{Everingham2009ThePV} and COCO datasets~\cite{lin2014microsoft}. 

Comparing MSCN with other methods in Table~\ref{tab:transfer-image-classification}. We observe similar results as Table~\ref{tab:linear_probe_siamese}, that MSCN improves SimCLR but performs worse with BYOL. We still suspect that the worse performance of BYOL could come from the non-optimal masking hyperparameters. Since most of the objects in Places-205 have a larger scale than the objects in iNaturalist 2018, the non-optimal masking hyperparameters cost more harm on the smaller scale objects than on larger-scale objects.

For the object detection and instance segmentation tasks, MSCN demonstrates superior performances over previous siamese network frameworks on VOC07+12 detection task and performs comparably to the state-of-the-art representation learning methods on
COCO dataset.

\begin{table}[h!]
    \centering
    \caption{Image classification transfer learning with a ResNet-50 pretrained on ImageNet-1K. We follow the standard evaluation protocol that trains the linear classifiers on fixed features with the same hyperparameters as other methods except for the learning rate.}
    \begin{tabular}{lll}
        \textbf{Method}  & \textbf{Place-205} &  \textbf{iNat18} \\
    \toprule
        SimCLR~\cite{chen2020simple} & 52.5 &  37.2 \\
        \rowcolor{LightCyan} \ \ \ \  +MSCN & 53.8 \textcolor{SeaGreen}{\small (+1.3)} & 38.2 \textcolor{SeaGreen}{\small (+1.0)} \\
        \midrule
        BYOL~\cite{grill2020byol} &  54.0 & 47.6  \\
        \rowcolor{LightCyan} \ \ \ \  +MSCN & 54.2 \textcolor{SeaGreen}{\small (+0.2)} & 44.3 \textcolor{RedViolet}{\small (-3.3)} \\
        % VICReg~\cite{grill2020byol} &  54.3 & 47.0  \\
        % \rowcolor{LightCyan} \ \ \ \  +MSCN & ? & ? \\
    \bottomrule
    \end{tabular}
    \label{tab:transfer-image-classification}
\end{table}

\begin{table}[h!]
    \centering
    \caption{Object detection and instance segmentation transfer learning with a ResNet-50 pretrained on ImageNet-1K. All VOC07+12 results using Faster R-CNN~\cite{ren2015faster} with C4 backbone variant~\cite{wu2019detectron2} finetuned 24K iterations. All COCO results using Mask R-CNN~\cite{he2017mask} with C4 backbone variant~\cite{wu2019detectron2} finetuned using the 1× schedule. 0.3 within the best are \underline{underlined}}
    \begin{tabular}{@{}lccccccccc@{}}
        \textbf{Method} & \multicolumn{3}{c}{\textbf{VOC07+12 det}} & \multicolumn{3}{c}{\textbf{COCO det}} & \multicolumn{3}{c}{\textbf{COCO instance seg}}\\
     \cmidrule(lr){2-4}\cmidrule(lr){5-7}\cmidrule(lr){8-10}
    & AP$_{\mathrm{all}}$ & AP$_{50}$ & AP$_{75}$ &  AP$^{\mathrm{bb}}$ & AP$^{\mathrm{bb}}_{50}$ & AP$^{\mathrm{bb}}_{75}$ & AP$^{\mathrm{mk}}$ & AP$^{\mathrm{mk}}_{50}$ & AP$^{\mathrm{mk}}_{75}$\\
    \toprule
    \textcolor{gray}{Supervised} & \textcolor{gray}{53.5} & \textcolor{gray}{81.3} & \textcolor{gray}{58.8} & \textcolor{gray}{38.2} & \textcolor{gray}{58.2} & \textcolor{gray}{41.2} & \textcolor{gray}{33.3} & \textcolor{gray}{54.7} & \textcolor{gray}{35.2} \\
    MoCo v2 \cite{He2020MomentumCF} & \underline{57.4} & 82.5 & 64.0 & \underline{39.3} & 58.9 & \underline{42.5} & \underline{34.4} & \underline{55.8} & \underline{36.5} \\
        SwAV~\cite{caron2020swav} & 56.1 & 82.6 & 62.7 & 38.4 & 58.6 & 41.3 & 33.8 & 55.2 & 35.9 \\
        SimSiam~\cite{chen2020simsiam} & 57 & 82.4 & 63.7 & \underline{39.2} & \underline{59.3} & 42.1 & \underline{34.4} & \underline{56.0} & \underline{36.7} \\
        Barlow Twins~\cite{zbontar2021barlow} & 56.8 & 82.6 & 63.4 & \underline{39.2} & \underline{59.0} & \underline{42.5} & \underline{34.3} & \underline{56.0} & \underline{36.5} \\
        
        \rowcolor{LightCyan} MSCN & \underline{57.5} & \underline{83.0} & \underline{64.4} & \underline{39.1} & \underline{59.1} & 42.1 & \underline{34.2} & \underline{55.7} & \underline{36.4} \\
        % \rowcolor{LightCyan} MSCN (w/ BYOL) & 57.1+ & 82.6+ & 64.0+ & 38.8+ & 58.9+ & 42.0+ & 33.9+ & 55.4+ & 36.1+ \\
    \bottomrule
    \end{tabular}
    \label{tab:transfer-object-detection}
\end{table}

\subsection{Ablation Study}

We conduct ablation experiments to gain insights into our masking design strategy. By default, we pretrain MSCN for 100 epochs with the masking design derived in Sec~\ref{sec:design}. We measure the performance by linear probe accuracy on ImageNet-1K.

\textbf{Masking Ratio} - We first explore the optimal masking ratio in Table~\ref{tab:ablation_mask_ratio}. A small masking ratio of $0.15$ is optimal for a ResNet-50 backbone. This matches the observation in \cite{Assran2022MaskedSN} that smaller networks prefer a smaller masking ratio. We also observe that the accuracy is relatively stable against the masking ratio up to $0.50$ with our masking strategy.

\begin{table}[h!]
    \centering
    \caption{Impact of masking ratio (fraction of randomly dropped grids in each random mask) during pretraining on ImageNet-1K linear probe accuracy (in \%). A small masking ratio is optimal for our masking strategy. The performance is stable up to a masking ratio of $0.50$ with our masking strategy.}
    \begin{tabular}{lcccccccc}
        \textbf{Masking Ratio} & 0.10 & 0.15 & 0.20 & 0.25 & 0.30 & 0.50 & 0.70 & 0.90 \\
        \toprule
        Accuracy (\%) & 67.5 & 67.6 & 67.5 & 67.4 & 67.4 & 66.8 & 66.2 & 62.6
    \end{tabular}
    \label{tab:ablation_mask_ratio}
\end{table}

\textbf{Masking Grid Size} - Mask grid size is an important hyperparameter that controls the balance between local and global features in the input. In masked siamese networks with ViTs, the masking grid size is fixed, and it is always set to match the patch boundaries. However, an optimal masking grid size can vary. In Table~\ref{tab:ablation_grid_size}, we show that learned representations can benefit from a better mask grid size. We observe a large grid size of s optimal for our current masking strategy with ConvNets backbone.

\begin{table}[h!]
    \centering
    \caption{Impact of masking grid size during pretraining on ImageNet-1K linear probe accuracy (in \%). We observe that a large grid size is optimal.}
    \begin{tabular}{lccccc}
        \textbf{Masking Grid Size} & 7 & 14 & 28 & 32 & 56 \\
        \toprule
        Accuracy & 67.1 & 67.2 & 67.4 & 67.4 & 67.2
    \end{tabular}
    \label{tab:ablation_grid_size}
\end{table}

\textbf{View Sharing} - In our masking strategy, we apply the standard augmentations to generate multiple views and then randomly apply masks on these views. One alternative is to apply random masks on the same augmented view. Table~\ref{tab:ablation_augmentation} shows that applying masks on the same view results in significantly worse representations.

As discussed in our design principle, the masks are used to prevent superficial solutions based on masked areas. It is still important to apply different augmentations to the original image to prevent the superficial solutions based on unmasked areas. 

\begin{table}[h!]
    \centering
    \caption{Impact of view-sharing during pretraining on ImageNet-1K linear probe accuracy (in \%). A view is generated with standard augmentations, including RandomResizedCrop, ColorJitter, HorizontalFlip, GaussianBlur, and Grayscale. Our standard approach applies masks to two different views. Here, we find that applying masks on a shared view results in significantly worse performance.}
    \begin{tabular}{lc}
        \textbf{Augmentation strategy} &  Accuracy \\
        \toprule
        Same view + random mask & 29.1 \\
        Different views + random mask & 65.6 \\
        \bottomrule
    \end{tabular}
    \label{tab:ablation_augmentation}
\end{table}

\section{Discussion and Future Directions}
\label{sec:discussion}

\textbf{Mask-only for Domain-knowledge Free Self-supervised Learning} - Human-designed augmentations leveraging domain knowledge are essential for siamese networks to learn useful representations. Similar to masked siamese networks~\cite{Assran2022MaskedSN} with ViTs, our approach combines standard augmentations with masking. However, it is desirable to find a domain-knowledge-free augmentation strategy so that this approach can be applied to more general domains or out-of-distribution scenarios. And it is intuitive to believe that the importance of domain-knowledge-specific augmentations will diminish with an increasing amount of pretraining data.

\textbf{Masked ConvNets for Generative Models} - Apart from siamese networks, denoising autoencoders~\cite{Vincent2008ExtractingAC, Vincent2010StackedDA} is another general approach for representation learning. Previous successful visual representation learning frameworks based on denoising autoencoders use global transformations such as rotation~\cite{Gidaris2018UnsupervisedRL} and Jigsaw~\cite{Goyal2019ScalingAB} and are outperformed by siamese networks. Recently, masked autoencoders~\cite{He2021MaskedAA} have demonstrated impressive performances in visual representation learning, only with ViTs. Unfortunately, masked autoencoders also fail to work with off-the-shelf ConvNets caused by similar problems as mentioned in Sec~\ref{sec:problem} for siamese networks. We suspect that our design in Masked Siamese ConvNets may also apply to masked autoencoders with ConvNets. We hope the discovery in this paper may shed light on general self-supervised learning and reduce the requirement for inductive bias of different architectures.

\section{Conclusion}
\label{sec:conclusion}

This work presents a method to add masking augmentation to siamese networks with ConvNets.
We first present the problems introduced by the use of masking as augmentation. We then carefully study how to gradually improve the downstream task performance by changing the masking strategy to solve or mitigate the problems.
Our method performs competitively on low-shot image classification benchmarks and outperforms previous methods on object detection benchmarks. We discuss several remaining issues and hope this work can provide useful data points for future general-purpose self-supervised learning.

\section*{Acknowledgement}

We thank Mahmoud Assran and Nicolas Ballas for kindly sharing insights from the Masked Siamese Networks paper. We thank Nicolas Carion for useful discussions on object detection. We thank Adrien Bardes for the useful discussion on the masking implementation and alternative approaches. We thank Quentin Garrido, Pascal Vincent, Randall Balestriero, Surya Ganguli, Yubei Chen, Florian Bordes for general discussions and feedbacks.

\bibliography{citation}
\bibliographystyle{plain}

\newpage
\appendix

\section{Implementation Detail}

\subsection{Pretraining}

We closely follow the original setting in \cite{chen2020simple} for our MSCN (w/ SimCLR) pretraining and original setting in \cite{grill2020byol} for our MSCN (w/ BYOL) pretraining.

\textbf{Augmentation} - For both methods, we use the same augmentation methods. Each augmented view is generated from a random set of augmentations from the same input image. 
We apply a series of standard augmentations for each view, including random cropping, resizing to 224x224, random horizontal flipping, a random color-jittering, randomly converting to grayscale, and a random Gaussian blur. These augmentations are applied symmetrically on two branches.

\textbf{Architecture} - 
For MSCN (w/ SimCLR), the encoder is a ResNet-50 network without the final classification layer followed by a projector. The projector is a two-layer MLP with input dimension 2048, hidden dimension 2048, and output dimension 256. The projector has ReLU between the two layers and batch normalization after every layer. This 256-dimensional embedding is fed to the infoNCE loss.
We use a temperature $0.2$ for the infoNCE loss.

For MSCN (w/ BYOL), the online encoder is a ResNet-50 network without the final classification layer. The online projector is a two-layer MLP with input dimension 2048, hidden dimension 4096, and output dimension 256. The predictor is a two-layer MLP with input dimension 256, hidden dimension 4096, and output dimension 256.
The projector and the predictor have ReLU between the two layers and batch normalization after every layer except the final linear layer.
The target encoder and projector are the exponential moving average of the online encoder and projector, with an initial momentum $\tau = 0.996$ with a cosine decay schedule to $1.0$ during the pretraining.

\textbf{Optimization} - 
We follow the training protocol in \cite{zbontar2021barlow} . For MSCN (w/ SimCLR), we use a LARS optimizer and a base learning rate 4.8 using cosine learning rate decay schedule. We pretrain the model for 800 epochs with 10 epochs warm-up with batch size 4096.

For MSCN (w/ BYOL), we use a LARS optimizer and a base learning rate 3.2 using cosine learning rate decay schedule. We pretrain the model for 1000 epochs with 10 epochs warm-up with batch size 4096.

\subsection{Linear Probe on ImageNet}

We closely follow the setting used in \cite{zbontar2021barlow} for our linear probe evaluation.
The linear classifier is trained for 100 epochs with a base learning rate of $1.0$ and a cosine learning rate schedule. We minimize the cross-entropy loss with the SGD optimizer with momentum and weight decay of $10^{-6}$. We use a batch size of 256. 

At training time, we use random resized crops to 224x224, followed by random horizontal flip and a High-pass filter. At test time, we resize the image to 256×256 and center-crop it to a size of 224×224 and followed by a High-pass filter.

\subsection{Finetuning}

We closely follow the setting used in \cite{zbontar2021barlow} for our finetuning evaluation.
We finetune the ResNet-50 encoder for 20 epochs with a learning rate of $0.002$ and the classifier with a base learning rate $0.5$. 
The
learning rate is multiplied by a factor of 0.2 after the 12th
and 16th epoch. We minimize the cross-entropy loss with
the SGD optimizer with momentum and do not use weight
decay. We use a batch size of 256. The image augmentations
are the same as in the linear evaluation setting.

\subsection{Object Detection}

We use the detectron2 library \cite{wu2019detectron2} for training the detection models. All the configuration files are from the VISSL library \cite{goyal2021vissl}, which are closely follow the evaluation settings from \cite{He2020MomentumCF}.
The backbone ResNet50 network for Faster R-CNN \cite{ren2015faster} and Mask
R-CNN \cite{he2017mask} is initialized using our pretrained model.

\textbf{VOC07+12} We use the VOC07+12 \cite{Everingham2009ThePV} trainval set of 16K
images for training a Faster R-CNN C-4
backbone for 24K iterations using a batch size of 16. The initial learning rate for
the model is 0.085 which is reduced by a factor of 10 after
18K and 22K iterations. We use linear warmup \cite{Goyal2019ScalingAB} with a slope of 0.333 for 1000 iterations.

\textbf{COCO} We train Mask R-CNN C-4 backbone on the COCO \cite{lin2014microsoft} 2017 train split for 90K iterations using a batch size of 16  The initial learning rate for
the model is 0.05 which is reduced by a factor of 10 after
60K and 80K iterations. We use linear warmup \cite{Goyal2019ScalingAB} with a slope of 0.333 for 1000 iterations. We report results on the 2017 val split.

\section{Additional Ablation Study}

We conduct additional ablation experiments to gain insights into our masking design strategy. By default, we pretrain MSCN with SimCLR loss for 100 epochs. We measure the performance by linear probe accuracy on ImageNet-1K.

\textbf{Focal Mask Probability} - 
 We explore the optimal focal mask probability in Table~\ref{tab:ablation_focal}.

\begin{table}[h!]
    \centering
    \caption{Impact of focal mask probability during pretraining on ImageNet-1K linear probe accuracy (in \%).}
    \begin{tabular}{lcccccc}
        \textbf{Focal Mask Probability} & 0.0 & 0.1 & 0.2 & 0.3 & 0.4 & 0.5 \\
        \toprule
        Accuracy & 67.0 & 67.0 & 67.4 & 67.5 & 67.6 & 67.5
    \end{tabular}
    \label{tab:ablation_focal}
\end{table}

\textbf{High-pass Sigma} -
We explore the optimal high-pass filter $\sigma$ in Table~\ref{tab:ablation_highpass}. In addition to the varying $\sigma$ for pretraining, we also update the high-pass filter $\sigma$ for the transformation during evaluation. In practice, we prefer a small $\sigma$ because there is less computational overhead.

\begin{table}[h!]
    \centering
    \caption{Impact of high-pass filter parameter $\sigma$ during pretraining on ImageNet-1K linear probe accuracy (in \%). }
    \begin{tabular}{lccccc}
        \textbf{High-pass Sigma} & 1.0 & 3.0 & 5.0 & 7.0 & 9.0 \\
        \toprule
        Accuracy & 66.7 & 67.3 & 67.4 & 67.4 &  67.3
    \end{tabular}
    \label{tab:ablation_highpass}
\end{table}

\textbf{Independent Mask Probability} -
 We explore the optimal independent mask probability in Table~\ref{tab:ablation_independent}.
 
\begin{table}[h!]
    \centering
    \caption{Impact of independent mask probability during pretraining on ImageNet-1K linear probe accuracy (in \%).}
    \begin{tabular}{lccccccc}
        \textbf{Independent Mask Probability} & 0.0 & 0.5 & 0.6 & 0.7 & 0.8 & 0.9 &1.0 \\
        \toprule
        Accuracy & 67.1 & 67.2 & 67.2 & 67.4 & 67.5 & 67.5 & 67.4    
        \end{tabular}
    \label{tab:ablation_independent}
\end{table}

\end{document}